# RULE BASED TRANSLITERATION SCHEME FOR ENGLISH TO PUNJABI

Deepti Bhalla[1], Nisheeth Joshi[2] and Iti Mathur[3]

[1,2,3]Apaji Institute, Banasthali University, Rajasthan, India

[1]deeptibhalla0600@gmail.com
[2]nisheeth.joshi@rediffmail.com
[3]mathur_iti@rediffmail.com

## ABSTRACT

*Machine Transliteration has come out to be an emerging and a very important research area in the field of machine translation. Transliteration basically aims to preserve the phonological structure of words. Proper transliteration of name entities plays a very significant role in improving the quality of machine translation. In this paper we are doing machine transliteration for English-Punjabi language pair using rule based approach. We have constructed some rules for syllabification. Syllabification is the process to extract or separate the syllable from the words. In this we are calculating the probabilities for name entities (Proper names and location). For those words which do not come under the category of name entities, separate probabilities are being calculated by using relative frequency through a statistical machine translation toolkit known as MOSES. Using these probabilities we are transliterating our input text from English to Punjabi.*

## 1. INTRODUCTION

Machine transliteration plays a very important role in cross-lingual information retrieval. The term, name entity is widely used in natural language processing. Name entity recognition has many applications in the field of natural language processing like information extraction i.e. extracting the structured text from unstructured text, data classification and question answering. In this paper we are only emphasizing on transliteration of proper names & locations using statistical approach. The process of transliterating a word from its native language to foreign language is called forward transliteration and the reverse process is called backward transliteration. This paper addresses the problem of forward transliteration i.e. from English to Punjabi. For Example, 'Silver' in English is transliterated to 'ਸਿਲਵਰ' into Punjabi instead of

'      '. Here our main aim is to develop a transliteration system which can be used to properly transliterate the name entities as well as the words which are not name entities from source language to target language. We have used English-Punjabi parallel corpus of proper names and locations to perform our experiment.

The remainder of this paper is organized as follows: In section 2 we described related work Section 3 gives a brief introduction about analysis of English and Punjabi script. Section 4 explains our methodology and experimental set up. Section 5 consists of evaluation and results. Finally in section 6 we conclude the paper.

International Journal on Natural Language Computing (IJNLC) Vol. 2, No.2, April 2013

## 2. RELATED WORK

A lot of work has been carried out in the field of machine transliteration. Kamal Deep and Goyal [1] have addressed the problem of transliterating Punjabi to English language using rule based approach. The proposed transliteration scheme uses grapheme based method to model the transliteration problem. This technique has demonstrated transliteration from Punjabi to English for common names and achieved accuracy of 93.22%. Sharma et al. [2] presented English-Hindi Transliteration using Statistical Machine Translation in different Notations. In this paper they have shown that transliteration in WX-notation gives improved results over UTF-notation by training the English-Hindi corpus (of Indian names) using Phrase based statistical machine translation. Kamal deep et al. [3] proposed Hybrid Approach for Transliteration from Punjabi to English. In this paper they addressed the problem of transliterating Punjabi to English language using statistical rule based approach. The Authors used letter to letter mapping as baseline and tried to find out the improvements by statistical methods.  kaur and Josan [4] proposed a framework which addresses the issue of Statistical Approach to Transliteration from English to Punjabi. In this Statistical Approach to transliteration is used from English to Punjabi using MOSES, a statistical machine translation tool. They applied transliteration rules and after that showed average percentage, accuracy of the system and BLEU score which comes out to be 63.31% and 0.4502 respectively. Padda et al. [5] presented conversion system on Punjabi Text to IPA. In this paper they described steps for converting Punjabi text to IPA. In this paper, they have shown how a letter of Punjabi maps directly to its corresponding IPA symbol which represents the sounds of spoken language.  Josan et al. [6] presented a novel approach to improve Punjabi to Hindi transliteration by combining a basic character to character mapping approach with rule based and Soundex based enhancements. Experimental results show that this approach effectively improves the word accuracy rate and average Levenshtein distance. Dhore et al. [7] have addressed the problem of machine transliteration where given a named entity in Hindi using Devanagari script need to be transliterated in English using CRF as a statistical probability tool and n-gram as a features set. In this approach, they presented machine transliteration of named entities for Hindi-English language pair using CRF as a statistical probability tool and n-gram as feature set. They achieved accuracy of 85.79% . Musa et al. [8] proposed Syllabification Algorithm based on Syllable Rules Matching for Malay Language. In this paper, authors presented a new syllabification algorithm for Malay language.

## 3. ANALYSIS OF ENGLISH AND PUNJABI SCRIPT

Punjabi is an ancient and Indo-Aryan language which is mainly used in regions of Punjab. It is one of the world's $14^{th}$ widely spoken languages. Its script is Gurumukhi which is based on Devanagri. Punjabi is one of the Indian language and official language of Punjab. Along with Punjab it is also spoken in neighbouring areas such as Haryana and Delhi. There are 19 Vowels (10 Independent Vowels and 9 dependent Vowels) and 41 Consonants in Punjabi language.
English on the other hand is based on Roman Script. It is one of the six international languages of United Nations. There are 26 letters in English. Out of which 21 are Consonants and 5 are Vowels.

## 4. METHODOLOGY AND EXPERIMENTAL SETUP

In our work, we have English Punjabi parallel corpus of name entities. We are basically emphasizing on transliteration of our input text with the help of rule matching. We are using statistical approach for transliteration that is we train and test our model using MOSES toolkit. With the help of GIZA++ we have calculated translation probabilities which are based on relative

68

frequency. For constructing our rules we are using syllabification approach. The syllable is a combination of vowel and consonant pairs such as V, VC, CV, VCC, CVC, CCVC and CVCC. Almost all the languages have VC, CV or CVC structures so we are using vowel and consonants as the basic phonification unit.

Some possible rules that we have constructed for syllabification are shown in table 1:

Table1: Possible combinations for syllables

| Syllable structure | Example | Syllabified form(English) | Syllabified form(Punjabi) |
|---|---|---|---|
| V | eka | [e] [ka] | [ ] [ ] |
| CV | tarun | [ta] [run] | [ ] [ ] |
| VC | angela | [an] [ge] [la] | [ ] [ ] [ ] |
| CVC | sidhima | [si] [dhi] [ma] | [ ] [ ] [ ] |
| CCVC | Odisha | [o] [di] [sha] | [ ] [ ] [ ] |
| VCC | obhika | [o] [bhi] [ka] | [ ] [ ] [ ] |

**V=Vowel, C=Consonant**

If we have a word P then it will be syllabified in the form of {p1,p2,p3….pn} where p1,p2..pn are the individual syllables. We have syllabified our English input using rule based approach for which we have used the following algorithm.

**Algorithm for syllable extraction**

1. Enter input string in English.

2. Identify Vowels and Consonants.

3. Identify Vowel-Consonant combination and consider it as one syllable.

4. Identify Consonants followed by Vowels and consider them as separate syllables.

5. Identify Vowels followed by two continuous Consonants as separate syllable.

6. Consider Vowel surrounded by two Consonants as separate syllable.

7. Transliterate each syllable into Punjabi.

International Journal on Natural Language Computing (IJNLC) Vol. 2, No.2, April 2013For converting our English input to equivalent Punjabi output we first recognize the name entities from our input sentence. For this we have used Stanford NER tool [10] which gives us name entities in English text. The text entered by the user is first analyzed and then pre-processed. After pre-processing if the selected input is a proper name or location then it is passed to the syllabification module through which the syllables are extracted. After selection of equivalent probability, syllables are combined to form the Punjabi word. If the selected input is not a name entity it will be passed to the syllabification module and will be transliterated with the help of probability matching. Separate probabilities are calculated for those words which are not name entities.

Table 2: Corpus

|  | English Words | Punjabi Words | English Sentences |
|---|---|---|---|
| Training | 25,500 | 25,500 |  |
| Testing | 6080 |  | 1500 |

Table 2 demonstrates the English-Punjabi parallel corpus that we have used for performing our experiment. In this we have trained our language model using IRSTLM toolkit [9]. Our transliteration system follows the steps which are represented in figure 1.

Example:

**Source sentence**: Teena is going to Haryana

**Target sentence**:

In this initially the name entities will be extracted which are Teena (proper name) and Haryana (location). After pre-processing, these words will be checked with the help of name entity recognizer for identifying whether they are name entities or not. If the word is a name entity of type proper name and location, they will be passed to the syllabification module. In this case the name **Teena** will be syllabified in the form of **Tee na** and location name **Haryana** will be syllabified in the form of **Har ya na.** After syllabification, these words will be checked for their corresponding English-Punjabi probabilities and equivalent Punjabi syllables are selected. The words other than the name entities are syllabified and matched against the separate probabilities. For example **going** are transliterated to **go ing**. Finally these syllables are joined to generate final output sentence.

Through language model training we have calculated N-gram probabilities which are based on relative frequency. For calculating the N-gram probability we are using the following formula [11].

$$P(w_n|w_{n-1}) = \frac{C(w_{n-1}|w_n)}{C(w_{n-1})} \quad (1)$$

Where probability of a word $w_n$ given a word $w_{n-1}$ is count ($w_{n-1}$, $w_n$)/count ($w_{n-1}$)

For example:

70

For a name **Kunal** probabilities for every syllable can be as follows:

$Prob(ੁ|ku) = \frac{C(ku,ੁ)}{C(ku)} = 0.8797654$

$prob(ਾਲ|nal) = \frac{C(nal,ਾਲ)}{C(nal)} = 0.7333333$

Final Score=Prob( |ku)×Prob( |nal)

=0.64516127

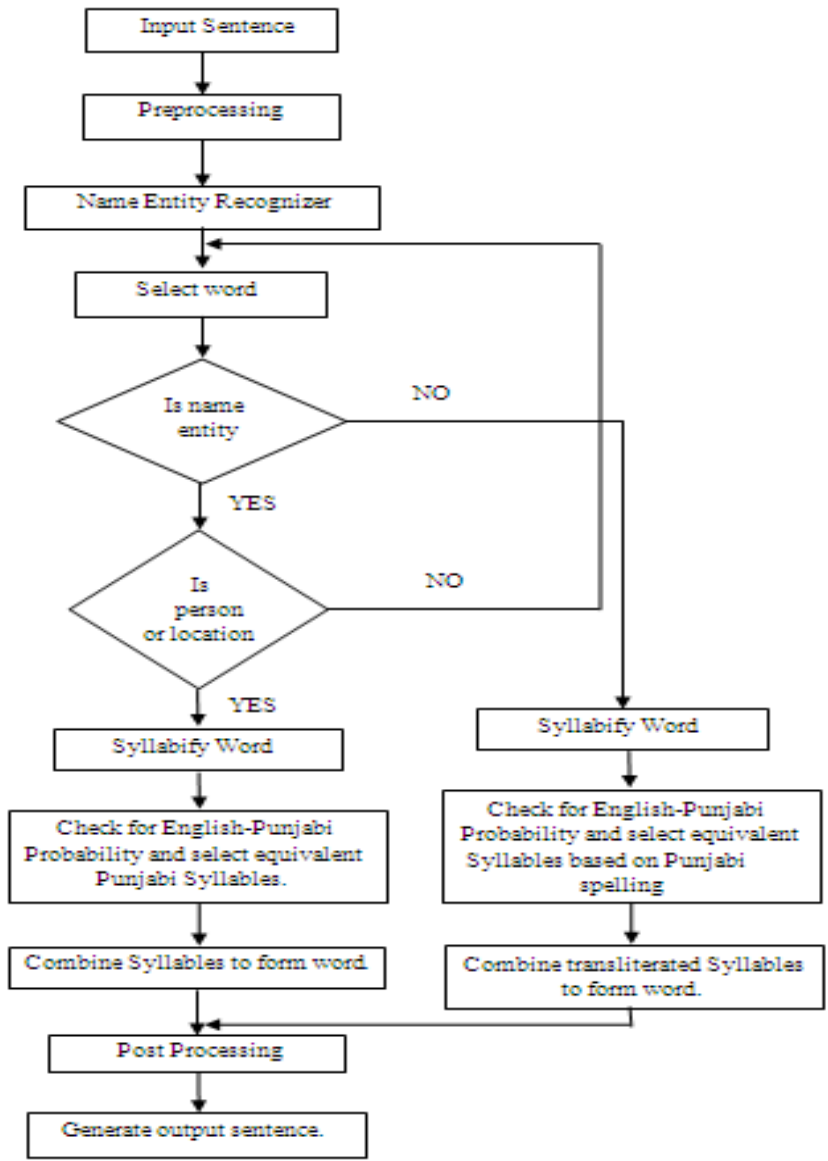

Figure 1: System architecture

After selecting the equivalent Punjabi syllable on the basis of probabilities these syllables are joined together and final output is obtained.

## 5. EVALUATION AND RESULTS

We have used a corpus of 1500 sentences which had 15982 words. Among them 45% were name entities i.e. 7192 were proper names, locations, organization and Date & Time. Among these 88% were name entities of type person & location i.e. 6328 were person names or location names. From these our system was able to correctly transliterate 6109 name entities. Remaining 55 % i.e. 8790 words were those that were not name entities and from these our system was able to transliterate 7987 words correctly. This provided us with accuracy of 88.19%. The accuracy was calculated using the formula:

$$Accuracy(\%) = \frac{correct\ Transliteration}{Total\ Name\ Entities} * 100 \qquad (2)$$

Some of the names which are correctly transliterated by our system are as follows:.

Table 3: Correct output

| | |
|---|---|
| spardha | |
| sanbosh | |
| kamaldeep | |
| haryana | |
| shelly | |

Some of the names which are not correctly transliterated by our system are as follows:

Table 4: Incorrect output

| | |
|---|---|
| falkanvai | |
| gowryharan | |
| seerajuddeen | |
| kolkandkar | |

## 6. CONCLUSION

In this paper we have performed our experiment using statistical machine translation tool. We have shown how we can extract syllables from our input text and perform transliteration of name entities (proper names and locations) and rest of the words other than name entities in source language to target language through probability calculation. This transliteration system will prove to be very beneficial. Through this approach we attained accuracy of 88.19%. In future we will work on remaining name entities that were not included in this experiment.

**Authors**

Deepti Bhalla is pursuing her M.Tech in Computer Science from Banasthali University, Rajasthan and is working as a Research Assistant in English-Indian Languages Machine Translation System Project sponsored by TDIL Programme, DEITY. She has her interest in Machine Translation specifically in English-Punjabi Language Pair. She has developed various tools on Punjabi Language Processing. Her current research interest includes Natural Language Processing and Machine Translation. 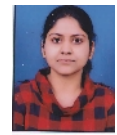

Nisheeth Joshi is a researcher working in the area of Machine Translation. He has been primarily working in design and development of evaluation Matrices in Indian languages. Besides this he is also actively involved in the development of MT engines for English to Indian Languages. He is one of the expert empanelled with TDIL programme, Department of electronics Information Technology (DEITY), Govt. of India, a premier organization which foresees Language Technology Funding and  Research in India. He has several publications in various journals and conferences and also serves on the Programme Committees and Editorial Boards of several conferences and journals. 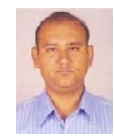

Iti Mathur is an assistant professor at Banasthali University. Her primary area of research is computational semantics and ontological engineering. Besides this she is also involved in the development of MT engines for English to Indian Languages. She is one of the experts empanelled with TDIL Programme, Department of Electronics Information Technology (DEITY), Govt. of India, a premier organization which foresees Language Technology Funding and Research in India. She has several publications in various journals and conferences and also serves on the  Programme Committees and Editorial Boards of several conferences and journals. 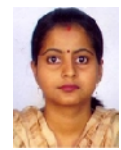